\documentclass{article}
\PassOptionsToPackage{numbers,compress}{natbib}
\usepackage[preprint]{neurips_2026}

\usepackage[utf8]{inputenc} 
\usepackage[T1]{fontenc}    
\usepackage[hidelinks]{hyperref}      
\usepackage{url}            
\usepackage{booktabs}       
\usepackage{amsfonts}       
\usepackage{nicefrac}       
\usepackage{microtype}      
\usepackage{xcolor}         
\usepackage{graphicx}
\usepackage{wrapfig}

\title{Atomic Task Graph: A Unified Framework for Agentic Planning and Execution}

\author{%
Yue Zhang$^{1}$\thanks{Equal contribution.} \quad
Sihan Chen$^{1}$\footnotemark[1] \quad
Ziwen Huang$^{1}$\footnotemark[1] \quad
Hanyun Cui$^{2}$ \quad
Kangye Ji$^{2}$ \quad
Zhi Wang$^{2}$\thanks{Corresponding author.} \\
$^{1}$South China University of Technology \\
$^{2}$Tsinghua Shenzhen International Graduate School, Tsinghua University
}

\begin{document}

\maketitle

\begin{abstract}
LLM-based agents have shown strong potential for solving complex multi-step tasks, yet existing performance improvements often rely on either scaling to larger backbone models or task-specific fine-tuning. The former incurs substantial computational costs, while the latter typically generalizes poorly across different tasks. Although prompt-based control is training-free and broadly applicable, existing methods still leaves input-output dependencies between subtasks implicit in textual trajectories, making verified intermediate results difficult to reuse. 
To address these limitations, we propose \textbf{A}tomic \textbf{T}ask \textbf{G}raph (\textbf{ATG}), which is a unified control framework across planninng and execution. 
Specifically, ATG maintains an explicit graph to expose dependencies and support reuse. During planninng, it recursively decomposes a high-level task into subtasks, forming a sequence of directed acyclic graphs (DAGs) whose evolution can be traced. During execution, the dependencies exposed by the ATG allow independent branches to be executed in parallel, thereby improving execution efficiency. When failures are detected, ATG leverages the graph evolution history to localize the error source and repair only the affected region, preserving validated regions unchanged.
Experiments show that ATG consistently outperforms strong baselines in success rate and execution efficiency across three interactive benchmarks using only 7B--8B backbones.
\end{abstract}

\section{Introduction}
LLM-powered autonomous agents have shown strong potential for complex multi-step problem solving by combining reasoning, planning, and external tool use \cite{survey1,survey2,schick2023toolformer,pal,gorilla,yao2023react,shinn2023reflexion}.
They have been applied to tasks such as web navigation\cite{webshop,webarena,mind2web}, embodied decision-making\cite{alfred,alfworld}, interactive question answering\cite{hotpotqa,webgpt}, and scientific reasoning\cite{scienceworld}. For multi-step tasks, however, performance depends not only on model capability but also on how planning and execution are organized\cite{tot,got,cotsc,zhou2023lats}. Scaling backbone models improves reasoning but incurs high training and inference costs\cite{palm,gpt3}, while task-specific fine-tuning is more efficient but often generalizes poorly across diverse tasks\cite{planpalm,flan}.

Existing prompting-based methods typically organize task solving as a sequential decision-making process\cite{yao2023react,shinn2023reflexion,cot,wang2023plan}. Even when some methods introduce trees or graphs, these structures are mostly used for candidate path exploration\cite{tot,got,pog}. They struggle to represent non-linear input-output dependencies among subtasks. The final executed solution still remains a linear textual trajectory. 

\begin{figure}[t]
    \centering
    \includegraphics[width=\linewidth]{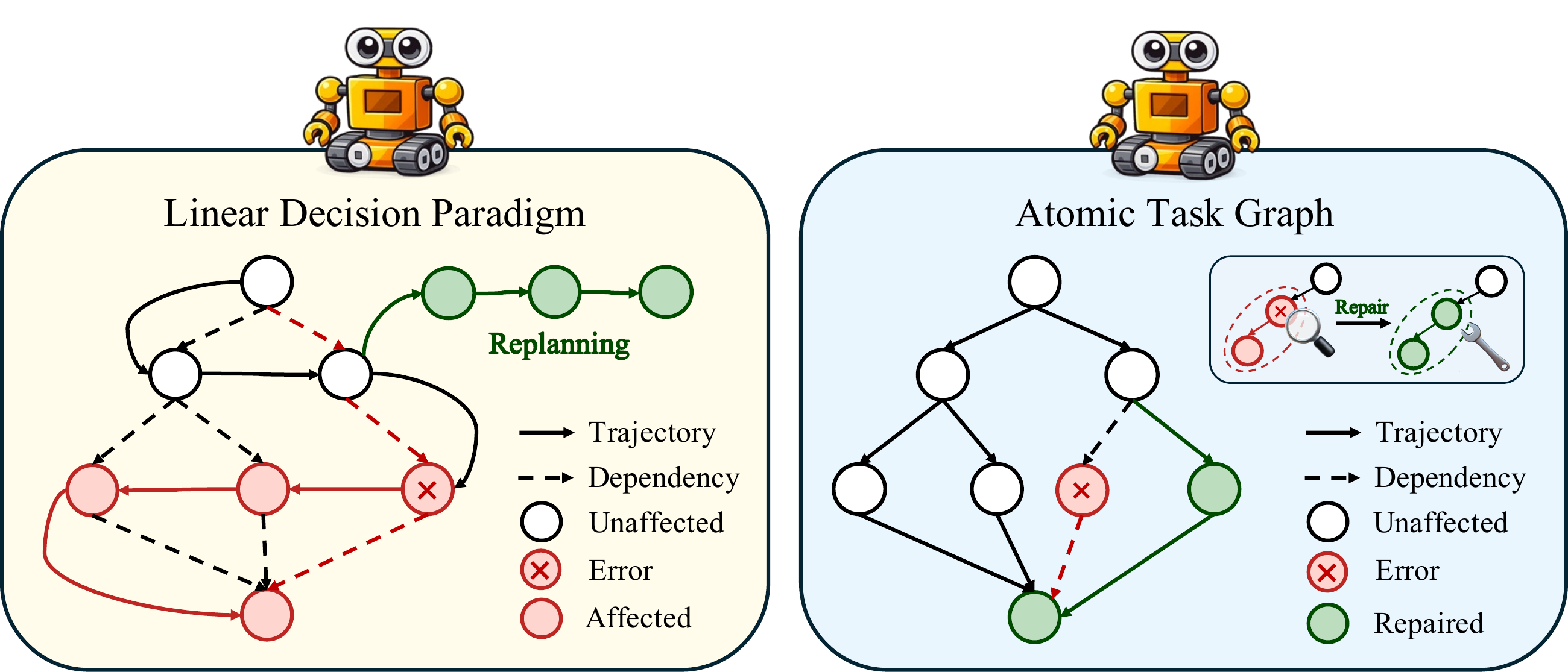}
    \caption{Comparison between the linear decision paradigm and Atomic Task Graph.}
    \label{fig:comparison}
\end{figure}

As shown in Figure~\ref{fig:comparison}, under linear decision paradigm, errors are coupled with previous steps and can propagate along the trajectory. Without an explicit dependency structure, failures cannot be localized to a clear affected region. As a result, existing repair methods often rely on continued correction, backtracking, or broad replanning, rather than precise local repair and reuse. Moreover, the growing textual context can further induce hallucinated actions in later execution stages.

\begin{figure}[t]
    \centering
    \includegraphics[width=\linewidth]{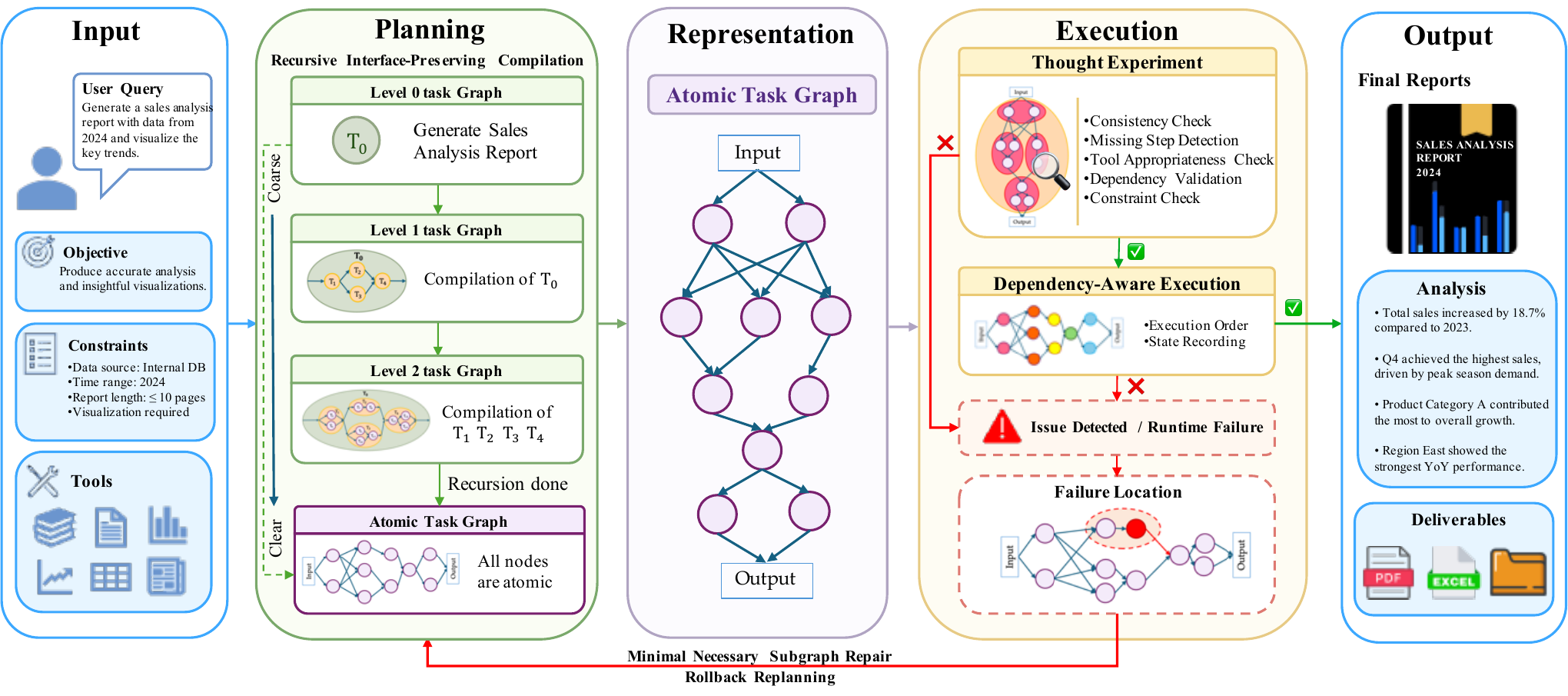}
    \caption{Visualization of framework.}
    \label{fig:framework}
\end{figure}

To address these limitations, we propose Atomic Task Graph (ATG), a unified control framework that integrates graph structure throughout planning and execution. Figure~\ref{fig:framework} illustrates the overall framework of ATG. Specifically, ATG formulates agent task solving as a DAG, explicitly exposing subtask dependencies, enabling parallel execution of independent branches, and localizing failures to the smallest affected region for reusing verified intermediate results.

Firstly, during planning, we aim to explicitly capture subtask dependencies in the plan by designing an interface-preserving recursive graph compilation method. This process refines a coarse user task into subgraphs until all nodes become atomic tool-use units, while preserving the input-output interface of each parent node at every step. The recursive compilation process naturally produces a sequence of graphs that record the evolution of the task from coarse to fine, which provides a structured basis for later tracing where and why an error occurs. As recursive decomposition proceeds, the context associated with each node becomes increasingly localized. Therefore, action generation no longer depends on growing textual context, reducing the risk of hallucinated actions. 

Secondly, during execution, to better leverage the explicit dependency structure, ATG designs dependency-aware graph execution and minimal necessary subgraph repair. Specifically, ATG allows independent branches to run in parallel for faster completion. It also performs a lightweight pre-execution thought experiment to detect potential failures. When a potential or runtime failure occurs, ATG freezes validated regions and repairs only the smallest affected subgraph, avoiding unnecessary global replanning.

We evaluate ATG on ALFWorld, WebShop, and ScienceWorld, which require long-horizon planning, interactive execution, state tracking, and error recovery. Across multiple open-source backbone models, ATG improves performance over representative baselines. Ablation studies further verify the importance of dependency-aware execution and localized graph repair.

Our key contributions are summarized as follows:

\begin{itemize}
    \item We propose \textbf{Atomic Task Graph (ATG)}, a unified control framework that represents LLM-agent task solving as a directed graph of atomic tool-use units and their input-output dependencies across agentic planning and execution.

    \item We introduce \textbf{interface-preserving recursive graph compilation}, which progressively exposes dependencies among subtasks. By recording graph evolution and localizing the context of each atomic node, this process supports later error tracing and reduces hallucinated actions caused by growing textual context.

    \item We design \textbf{dependency-aware graph execution with minimal necessary subgraph repair}, enabling parallel execution of independent branches, failure localization, and localized repair without unnecessary global replanning.
\end{itemize}

\section{Related Work}
\paragraph{LLM Agents.}
LLMs have emerged as a promising foundation for building autonomous agents, enabling systems to perform complex tasks through reasoning, planning, tool use, memory, and interaction with external environments\cite{survey1,survey2}. Existing studies extend LLMs from passive text generators to interactive decision-making agents in diverse scenarios, such as web navigation\cite{webshop,webarena,mind2web}, embodied task solving\cite{alfred,alfworld}, tool-augmented reasoning\cite{schick2023toolformer,pal,gorilla,hugginggpt}, interactive question answering\cite{hotpotqa,webgpt}, software engineering\cite{sweagent,swebench}, and scientific discovery\cite{scienceworld,chem,cos}. These agents typically follow a perception-reasoning-action loop, where the model observes task states, reasons over historical context, selects actions, and updates its behavior according to environmental feedback. Despite their progress, long-horizon agent tasks still require reliable organization of intermediate steps, dependencies, execution states, and failure recovery. Therefore, beyond improving model capability itself, a central challenge is how to design effective control frameworks for agentic planning and execution. 

\paragraph{Agentic Control Framework.}
Existing LLM-based agent control frameworks can be broadly divided into prompt-based and fine-tuning-based approaches. Prompt-based methods steer agents through natural-language interaction patterns, including implicit textual trajectories\cite{yao2023react,shinn2023reflexion,cotsc,cot}, explicit planning structures\cite{tot,got,zhou2023lats,pog}, and multi-agent coordination\cite{li2023camel,hong2024metagpt,wu2024autogen,yu2025dyntaskmas,llmdebate,multipersona}. These methods are training-free and flexible, but their control states are often embedded in growing contexts, planning-time search structures, or multi-agent conversations. Fine-tuning-based methods improve agentic capabilities by adapting model parameters with demonstrations, interaction traces or feedback signals\cite{qin2023toolllm,agentun,fireact}. While effective for improving domain-specific tool use and execution behavior, they require additional training cost and may generalize poorly to unseen tasks, tools, or environments.

\section{Problem Formulation}
\paragraph{Definition 1: Task.}
Let a task be denoted by $x\in\mathcal{X}$, where $x$ is a concrete user-provided problem with a specified objective, constraints, and desired output. For task $x$, let $\mathcal{Y}_x$ denote the set of acceptable final outputs. The goal is to produce some $y\in\mathcal{Y}_x$. For complex tasks, such an output is usually not obtained in a single step, but through multiple intermediate operations.

\paragraph{Definition 2: Tools.}
We define the tool space as $\mathcal{T}=\{f_k:\mathcal{I}_{f_k}\rightarrow\mathcal{O}_{f_k}\mid k=1,2,\dots,|\mathcal{T}|\}$, where each tool $f_k$ maps an input from $\mathcal{I}_{f_k}$ to an output in $\mathcal{O}_{f_k}$. In this paper, each tool is treated as an atomic functional unit, whose internal process is not further decomposed and which participates in task solving only through its input-output interface. A tool may be a search API, a calculator, a code executor, a database query module, or another task-specific function.

\paragraph{Definition 3: Agent.}
Given a task $x$ and a tool space $\mathcal{T}$, we define an agent as an intelligent entity that can formulate an appropriate plan, and use available tools to solve complex problems. During planning, the agent represents its plan as an explicit DAG $G=(V,E)$, where each node $v_j=(i_j,f_j,o_j)$ corresponds to one concrete tool call, where $f_j\in\mathcal{T}$ is the selected tool, $i_j$ is its input, and $o_j$ is its output. Each edge $e_{jk}: v_j\rightarrow v_k$ indicates that the output $o_j$ of node $v_j$ is used as part of the input $i_k$ of node $v_k$, thereby explicitly representing the dependency between operations. This formulation is general: linear chains and tree-structured plans can both be viewed as special cases of DAGs.
During execution, the agent uses tools strictly according to the plan specified by the DAG to complete the task.

\paragraph{Core Problem.}
Given task $x$ and tool set $\mathcal{T}$, the goal is to find a feasible DAG $G^\star=(V^\star,E^\star)$ such that (1) each node is a valid tool invocation, (2) each edge represents a correct input-output dependency, and (3) executing the DAG in a topological order yields an output $y\in\mathcal{Y}_x$. In other words, the essence of agent problem solving is to search within the tool space for a valid combination and execution order of tools that transforms the initial task input into a final acceptable output. For complex tasks, this process naturally decomposes the task into subtasks, solves each subtask with suitable tools, and then integrates the intermediate results into a complete answer.

\section{Method}
As shown in Figure~\ref{fig:framework}, ATG consists of three key stages. 
First, \textit{Interface-Preserving Recursive Graph Compilation} recursively refines a coarse task into an atomic task graph while preserving each node's external input-output interface (\S4.1). 
Second, \textit{Dependency-Aware Execution} executes atomic nodes according to explicit graph dependencies, enabling parallel scheduling and state tracking (\S4.2). 
Third, \textit{Minimal Necessary Subgraph Repair} localizes failures through the refinement history and repairs only the affected subgraph while freezing validated components (\S4.3).

\subsection{Interface-Preserving Recursive Graph Compilation}
\begin{figure}[t]
    \centering
    \includegraphics[width=\linewidth]{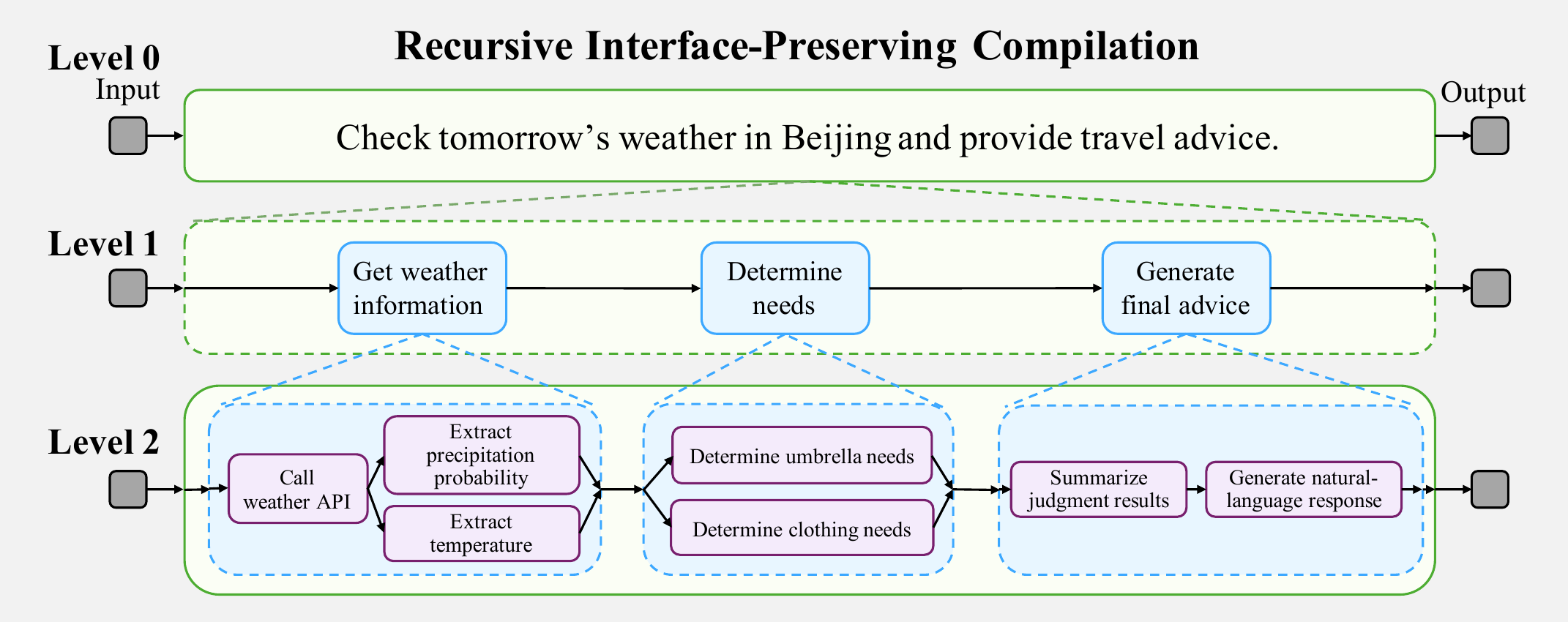}
    \caption{This figure shows how ATG recursively compiles a coarse task into atomic steps. The task “check tomorrow’s weather in Beijing and provide travel advice” is first split into weather retrieval, need determination, and advice generation, then further refined into executable steps such as calling the weather API, extracting weather features, judging umbrella and clothing needs, and generating the final response. Each refined subgraph preserves the parent node’s input-output interface.}
    \label{fig:Compilation}
\end{figure}
Complex user tasks are often too coarse for direct tool execution. Existing decomposition methods can split them into subtasks, but often lack explicit structures for error localization, subgraph reuse, and localized repair, while long textual histories may induce hallucinated actions. To address these issues, we formulate planning as a recursive graph compilation process.

\paragraph{Recursive Graph Compilation.}
Starting from an initial coarse-grained task graph, we recursively refine non-atomic nodes until every node becomes an atomic tool-use unit. At each refinement step, the LLM is only allowed to access the historical context directly relevant to the current node. A parent node $v$ is replaced by a subgraph $G_v$, whose internal nodes provide a more fine-grained decomposition of the subproblem represented by $v$. As the recursion depth increases, the context window attended to by each node becomes progressively narrower.

\paragraph{Interface preservation.}
The key design principle is that refinement must preserve the semantic interface of the parent node. Specifically, if the parent node $v$ is associated with an external input interface $i_v$ and output interface $o_v$, then the compiled subgraph $G_v$ must collectively consume the same external inputs and produce an output compatible with $o_v$. As a result, replacing $v$ with $G_v$ does not change how the rest of the graph interacts with that computation. This interface-preserving property makes recursive refinement compositional: each node can be expanded into a subgraph, while the surrounding graph remains structurally stable.

\paragraph{Termination.}
The recursion stops when each node corresponds to a single atomic tool-use unit that can be directly executed. The resulting graph is the final compiled atomic task graph. In this sense, planning is not merely the generation of a one-shot plan, but a progressive compilation process from coarse semantic intent to executable graph structure.

\paragraph{Refinement history.}
Importantly, we record the intermediate graph obtained at every refinement round. This produces a sequence of graphs with gradually increasing structural clarity and decreasing task granularity. The refinement history serves two roles. First, it provides an interpretable trace of how abstract task nodes evolve into executable units. Second, it becomes the basis for localized self-correction: when an error is later detected, the system can trace the problematic node back to its earlier refinement stages and revise only the relevant substructure rather than restarting from scratch.

\subsection{Dependency-Aware Execution}
Once the graph is compiled, execution proceeds directly over its explicit dependencies. Unlike loop-style agents that repeatedly infer the next action from a growing textual trajectory, ATG executes nodes through dependency-aware scheduling, validation, and state tracking on the graph.

\paragraph{Thought experiment.}
The thought experiment serves as a pre-execution filter that moves part of the recovery burden from expensive real-world execution to cheaper internal simulation. This step simulates the intended execution process internally, with the goal of exposing latent structural or operational failures before incurring real execution cost.
Concretely, the system checks whether the current decomposition, tool assignment, dependency structure, and node interfaces are likely to support successful execution. This validation may reveal several classes of issues, including incorrect tool selection, missing intermediate steps, invalid dependency assumptions, interface mismatches between connected nodes, or implausible execution paths. During this process, we record the step at which a failure is exposed, together with the associated error type and diagnostic information.

\paragraph{Execution order.}
Nodes are executed according to the topological order induced by graph dependencies. A node becomes executable when all of its predecessor nodes have finished and its required inputs have been resolved. Thus, execution readiness is determined by the graph itself, rather than requiring the agent to repeatedly reason over the entire history.
Whenever multiple nodes have no unresolved dependency conflicts, they can be executed in parallel. This allows the system to exploit structural parallelism exposed by the graph, reducing latency for multi-step tasks with partially independent branches.

\paragraph{State recording.}
During execution, we record each node's input, output, execution status, and any error messages produced by the corresponding tool call. This node-level state tracking is essential for downstream recovery, because it provides precise evidence for localizing which region of the graph failed and which previously completed regions can be safely reused.

\subsection{Minimal Necessary Subgraph Repair}

Failures in tool-assisted multi-step tasks are often local rather than global. A mistake may arise from one incorrect tool choice, one missing intermediate node, or one invalid dependency assumption, while many other parts of the graph remain correct. Replanning the entire task in such cases is unnecessarily expensive and may destroy already validated structure. We therefore introduce a minimal necessary subgraph repair mechanism.

\paragraph{Failure localization.}
When a failure occurs, either during the thought experiment or real execution, the system first localizes it to a failed atomic node $v_f$, or a small set of failed atomic nodes $\mathcal{F}$. These nodes are then traced back through the graph evolution history produced by recursive compilation. When multiple failed nodes are involved, we identify their lowest common historical ancestor $a_f$, which is the smallest ancestor node from which the failed region was derived. This ancestor marks the original planning scope where the failure was introduced and provides a natural boundary for localized repair.

\paragraph{Minimal Necessary Subgraph Repair.}
Given the failed node and its lowest common historical ancestor, the system constructs a minimal repair subgraph covering the failed node, its relevant upstream context, and downstream nodes affected by its output. The remaining ATG is frozen to preserve validated states and avoid redundant recomputation. The system then repairs only this subgraph by replacing incorrect tools, inserting missing nodes, or adjusting local dependencies, and reintegrates it into the main graph while preserving its external input-output interface.

\section{Experiments}

\subsection{Experimental Settings}
\label{exp}
\paragraph{Datasets and Metrics.}
We evaluate our method on three long-horizon interactive agent benchmarks: \textbf{ALFWorld}\cite{alfworld}, a text-based embodied benchmark for multi-step household task completion; \textbf{WebShop}\cite{webshop}, a simulated e-commerce benchmark for product search, comparison, and purchase decisions; and \textbf{ScienceWorld}\cite{scienceworld}, a text-interactive benchmark for multi-step scientific reasoning and exploration. ALFWorld uses a binary reward of 0 or 1 to indicate task success, while WebShop and ScienceWorld provide dense rewards from 0 to 1 to measure task completion. For all datasets, we use the average reward over evaluation tasks as the final metric.

\paragraph{Models and Baselines.}
We instantiate our framework on three representative open-source backbone models: \textbf{Mistral-7B-Instruct-v0.2}\cite{mistral7b}, \textbf{Gemma-1.1-7B-it}\cite{gemma}, and \textbf{Meta-Llama-3-8B-Instruct}\cite{llama3}. To compare different control paradigms, we evaluate against five baselines: \textbf{ReAct}\cite{yao2023react} and \textbf{Reflexion}\cite{shinn2023reflexion} for implicit control, \textbf{Tree-of-Thoughts (ToT)}\cite{tot} and \textbf{Plan-over-Graph (PoG)}\cite{pog} for explicit control, and \textbf{CAMEL}\cite{li2023camel} for multi-agent control. We also use \textbf{GPT-3.5-Turbo} and \textbf{GPT-4}\cite{gpt4} with ReAct as strong large-model baselines. 

\paragraph{Implementation and Evaluation Protocol.}
ATG operates purely at inference time: it does not require task-specific fine-tuning, additional supervision, external demonstrations, or any parameter updates to the backbone model. 
For each open-source backbone, ATG and all backbone-dependent baselines share the same checkpoint, tokenizer, decoding configuration, action budget, and environment interface. 
This ensures that performance differences mainly reflect the effect of the control framework rather than differences in model capacity or evaluation conditions. 
Unless otherwise specified, we aggregate seen and unseen evaluation instances and report the overall average performance over the full test set. 
All experiments are conducted on 6 NVIDIA RTX 4090 GPUs.

\subsection{Results}
\paragraph{Main Results.}
As shown in Table~\ref{tab:main_results}, ATG consistently achieves the best performance across all three long-horizon agent benchmarks and all open-source backbone models. Compared with implicit-control baselines such as ReAct and Reflexion, ATG brings substantial gains, showing that maintaining planning and execution only through linear textual trajectories is insufficient for reliable long-horizon task solving. For example, on Mistral-7B, ATG improves over ReAct by 49.16 points on ALFWorld and 48.12 points on WebShop, demonstrating a clear advantage in embodied interaction and web decision-making tasks. Explicit-control and multi-agent baselines, including CAMEL, ToT, and PoG, improve over implicit methods but still remain behind ATG. Taking PoG as the strongest baseline on Mistral-7B, ATG surpasses it by 32.01 points on ALFWorld and 38.57 points on WebShop, suggesting that simply introducing planning structures such as trees, graphs, or agent roles is not enough. The key difference is that ATG uses the graph as an executable substrate for dependency tracking, scheduling, and failure recovery. Moreover, ATG shows stable improvements across different backbones, with stronger backbone models further improving the final performance. Notably, ATG with Llama-3-8B surpasses GPT-4 ReAct on ALFWorld and WebShop, while all ATG variants substantially outperform GPT-3.5-Turbo ReAct. This indicates that effective control mechanisms can significantly narrow the gap between smaller open-source models and larger proprietary models without relying solely on backbone scaling.

\begin{table*}[t]
\centering
\caption{Main results on ALFWorld, WebShop, and ScienceWorld under different backbone models. We report average task-level performance on each benchmark. Higher is better.}
\label{tab:main_results}
\small
\setlength{\tabcolsep}{8pt}
\renewcommand{\arraystretch}{1.2}
\begin{tabular}{c|c|ccc}
\toprule
\textbf{Backbone} & \textbf{Method} & \textbf{ALFWorld} & \textbf{WebShop} & \textbf{ScienceWorld} \\
\midrule

GPT-3.5-Turbo & ReAct        & 7.30  & 43.57 & 14.67 \\
GPT-4         & ReAct        & 41.24 & 64.34 & 66.16 \\
\midrule

              & ReAct        & 6.57  & 14.63 & 19.12 \\
              & Reflexion    & 8.84  & 16.64 & 19.53 \\
Mistral-7B    & CAMEL        & 11.68 & 18.05 & 22.37 \\
              & ToT          & 18.25 & 20.31 & 25.44 \\
              & PoG          & 23.72 & 24.18 & 28.63 \\
\cline{2-5}
              & \textbf{ATG} & \textbf{55.73} & \textbf{62.75} & \textbf{49.81} \\
\midrule

              & ReAct        & 4.38  & 5.93  & 3.54  \\
              & Reflexion    & 5.11  & 7.71  & 4.41  \\
Gemma-7B      & CAMEL        & 7.66  & 9.48  & 7.92  \\
              & ToT          & 25.91 & 10.84 & 11.32 \\
              & PoG          & 32.85 & 12.56 & 13.85 \\
\cline{2-5}
              & \textbf{ATG} & \textbf{58.71} & \textbf{64.93} & \textbf{52.03} \\
\midrule

              & ReAct        & 3.29  & 19.32 & 23.67 \\
              & Reflexion    & 4.38  & 22.73 & 26.28 \\
Llama-3-8B    & CAMEL        & 8.76  & 24.85 & 28.10 \\
              & ToT          & 15.33 & 27.40 & 31.25 \\
              & PoG          & 21.17 & 31.66 & 35.72 \\
\cline{2-5}
              & \textbf{ATG} & \textbf{63.65} & \textbf{68.36} & \textbf{56.79} \\
\bottomrule
\end{tabular}
\end{table*}

\begin{figure*}[t]
    \centering
    \includegraphics[width=\textwidth]{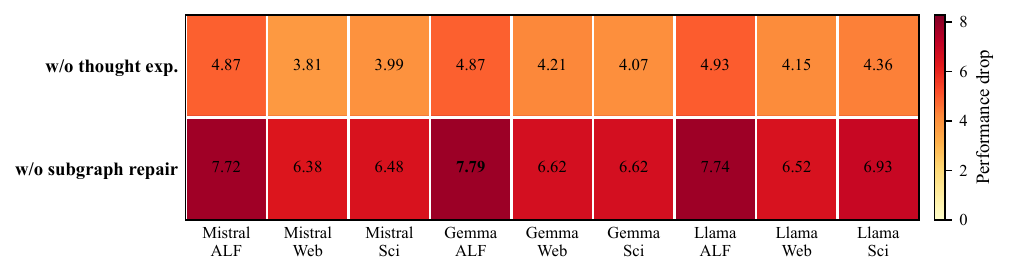}
    \caption{Ablation results of key ATG components across different backbone models. We report average task-level performance on each benchmark. Higher is better.}
    \label{fig:ablation}
\end{figure*}

\paragraph{Approach Ablations.}
As shown in Figure~\ref{fig:ablation}, both the pre-execution thought experiment and minimal necessary subgraph repair are important to ATG. Removing the thought experiment leads to consistent performance drops across all backbones, indicating that pre-execution checking helps identify invalid dependencies, missing steps, and likely planning failures before costly environment interaction. For instance, on Mistral-7B, removing this module reduces performance by 4.87 points on ALFWorld and 3.99 points on ScienceWorld. Removing minimal necessary subgraph repair causes a larger degradation in most cases; on Mistral-7B, the drop reaches 7.72 points on ALFWorld and 6.48 points on ScienceWorld, and similar trends are observed on Llama-3-8B. This suggests that local repair plays a more direct role in long-horizon tasks, where failures often originate from a small subset of erroneous nodes or dependencies rather than the entire plan. By preserving validated regions and repairing only the affected subgraph, ATG avoids unnecessary global replanning and improves execution stability. Overall, these ablations show that the thought experiment and subgraph repair are complementary: the former improves the reliability of the initial executable graph, while the latter enhances robustness during execution. Therefore, ATG’s gains come not merely from representing tasks as graphs, but from using the graph as a structured substrate for validation, execution, and recovery.

\begin{table*}[t]
\centering
\begin{minipage}{0.58\linewidth}
\centering
\caption{Average Steps. For ATG, parallel branches enabled by dependency-aware execution are counted as one step. Lower is better.}
\label{tab:avg_steps}
\small
\setlength{\tabcolsep}{5pt}
\renewcommand{\arraystretch}{1.15}
\begin{tabular}{l|c|c|c}
\toprule
\textbf{Method} & \textbf{ALFWorld} & \textbf{WebShop} & \textbf{ScienceWorld} \\
\midrule
ReAct      & 31.42 & 8.76 & 47.35 \\
Reflexion  & 29.18 & 8.21 & 44.62 \\
CAMEL      & 27.64 & 7.93 & 42.17 \\
ToT        & 25.83 & 7.48 & 39.86 \\
PoG        & 24.57 & 7.12 & 37.94 \\
\textbf{ATG} & \textbf{18.36} & \textbf{5.84} & \textbf{29.72} \\
\bottomrule
\end{tabular}
\end{minipage}
\hfill
\begin{minipage}{0.38\linewidth}
\centering
\caption{Hallucinatory Action Rates. Proportion of trajectories containing invalid or hallucinated actions. Lower is better.}
\label{tab:hallucination}
\small
\setlength{\tabcolsep}{6pt}
\renewcommand{\arraystretch}{1.15}
\begin{tabular}{l|c}
\toprule
\textbf{Method} & \textbf{ALFWorld} \\
\midrule
ReAct      & 42.86\% \\
Reflexion  & 38.57\% \\
CAMEL      & 35.71\% \\
ToT        & 31.43\% \\
PoG        & 28.57\% \\
\textbf{ATG} & \textbf{12.14\%} \\
\bottomrule
\end{tabular}
\end{minipage}
\end{table*}

\subsection{Analysis}

\paragraph{Dependency-aware execution reduces execution steps.}
\begin{wrapfigure}{r}{0.48\linewidth}
    \vspace{-1.0em}
    \centering
    \includegraphics[width=\linewidth]{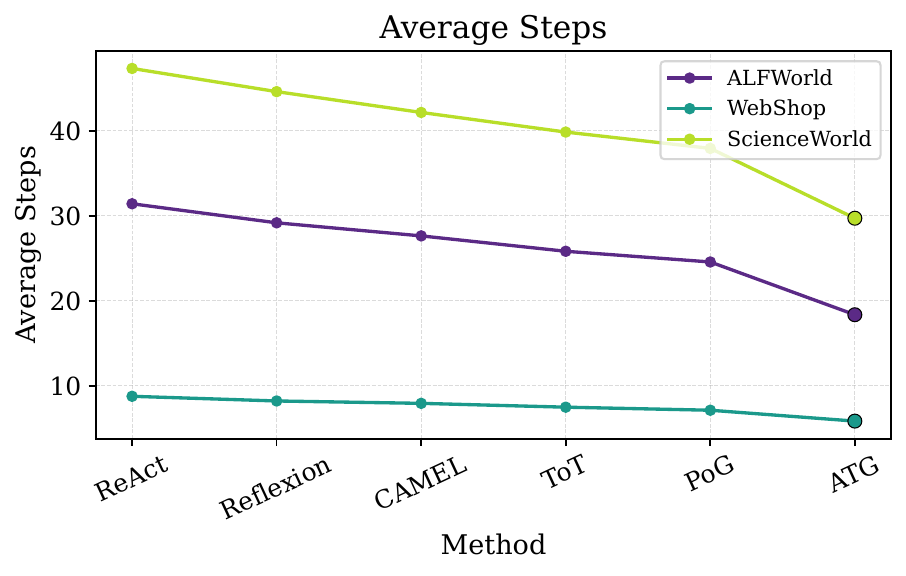}
    \caption{Average execution steps across three benchmarks.Lower is better.}
    \label{fig:avg_steps}
    \vspace{-1.2em}
\end{wrapfigure}
We first analyze whether the explicit dependency structure of ATG improves execution efficiency. 
As shown in Figure~\ref{fig:avg_steps}, ATG consistently requires fewer execution steps than all baselines across the three benchmarks. 
Compared with ReAct, ATG reduces the average number of steps from 31.42 to 18.36 on ALFWorld, from 8.76 to 5.84 on WebShop, and from 47.35 to 29.72 on ScienceWorld. 
Even compared with PoG, the strongest structured-planning baseline in terms of step efficiency, ATG still reduces the steps by 25.3\%, 18.0\%, and 21.7\% on ALFWorld, WebShop, and ScienceWorld, respectively. 
This indicates that the gain of ATG does not merely come from introducing a graph-like planning structure, but from using the graph as an executable dependency substrate. 
By explicitly identifying independent branches and executing them in parallel, ATG avoids unnecessary sequential decision making and reduces the effective execution depth of long-horizon tasks.

\paragraph{ATG reduces hallucinatory actions.}
We further examine whether ATG can reduce invalid or hallucinated actions during interaction. 
Since ALFWorld provides explicit feedback for invalid actions, we report the proportion of trajectories containing invalid or hallucinated actions in Table~\ref{tab:hallucination}. 
Prompt-based baselines suffer from relatively high hallucinatory action rates, with ReAct reaching 42.86\% and Reflexion reaching 38.57\%. 
Although explicit-control baselines such as ToT and PoG reduce this rate to some extent, they still produce hallucinated actions in 31.43\% and 28.57\% of trajectories, respectively. 
In contrast, ATG reduces the rate to 12.14\%, corresponding to a 71.7\% relative reduction over ReAct and a 57.5\% relative reduction over PoG. 
This result supports our claim that linear textual trajectories can accumulate irrelevant context and induce hallucinated actions in later stages, while ATG localizes the context of each atomic node and performs pre-execution validation before interacting with the environment. 
Therefore, ATG improves not only task success, but also the reliability of agent actions.

\begin{figure*}[t]
    \centering
    \includegraphics[width=\linewidth]{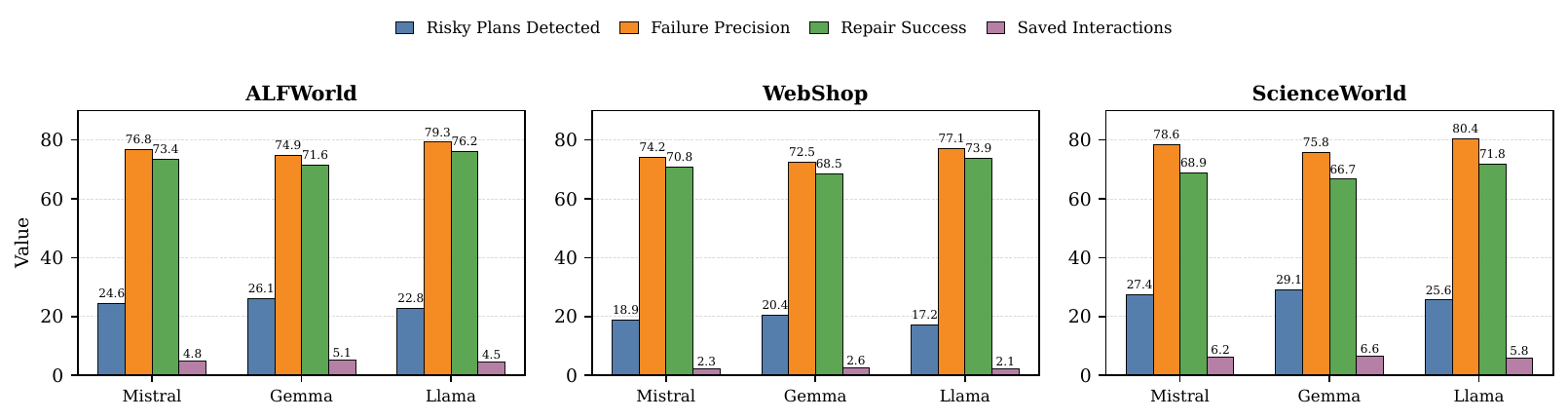}
    \caption{Effectiveness of pre-execution thought experiment across different backbone models. We report risky plans detected before execution, failure precision, repair success rate, and saved environment interactions.}
    \label{fig:thought_experiment}
\end{figure*}
\paragraph{Pre-execution thought experiment catches failures early.}
We also examine the effect of the pre-execution thought experiment. 
Rather than directly executing a compiled graph in the environment, ATG first performs a lightweight internal validation to identify potentially risky plans. 
As shown in Figure~\ref{fig:thought_experiment}, this step detects 24.6\%, 18.9\%, and 27.4\% risky plans on ALFWorld, WebShop, and ScienceWorld under Mistral-7B, respectively. 
The detected failures are also reliable, with failure precision consistently above 74\% across most settings. 
This suggests that the thought experiment can effectively expose missing steps, invalid dependencies, tool mismatches, and implausible execution paths before real environment interaction. 
By repairing risky plans early, ATG reduces wasted interactions and prevents local planning errors from developing into cascading runtime failures.

\paragraph{ATG scales consistently across backbones.}
\begin{wrapfigure}{r}{0.45\linewidth}
    \vspace{-1.0em}
    \centering
    \includegraphics[width=0.8\linewidth]{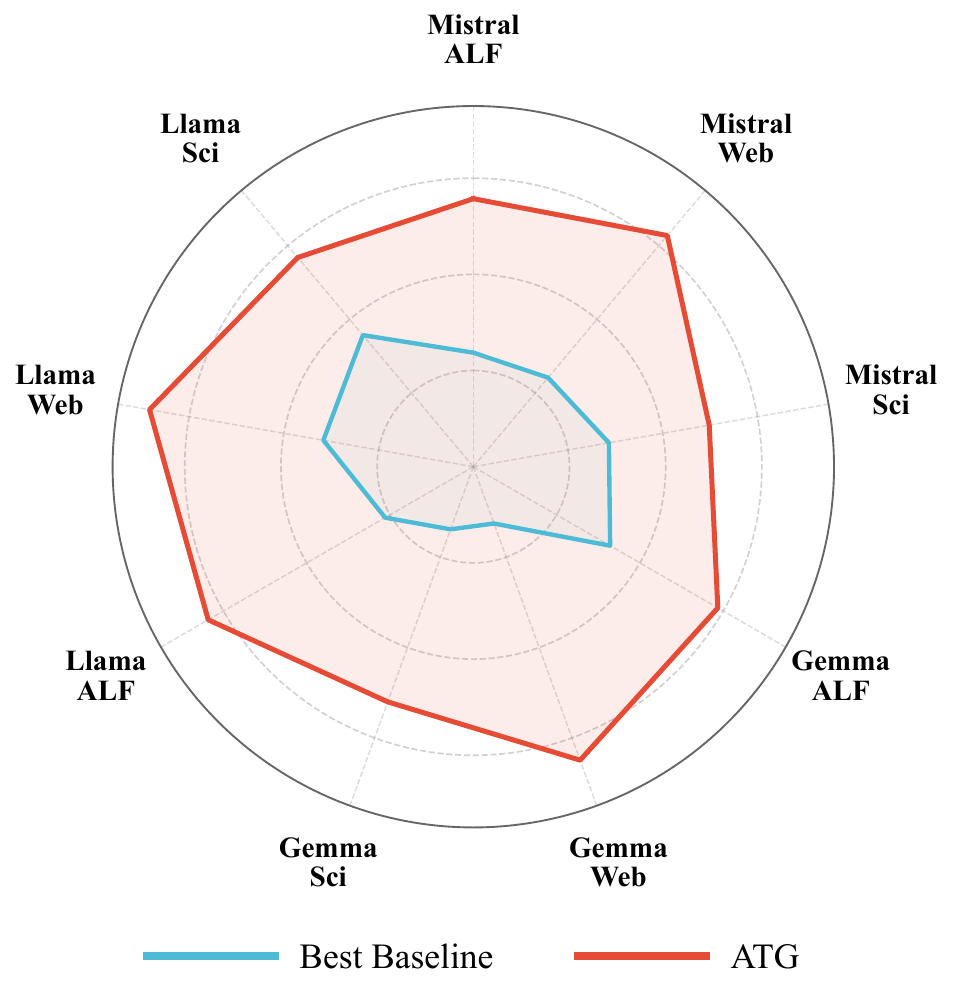}
    \caption{Generalization of ATG across different open-source backbone models. }
    \label{fig:atg_backbone_generalization}
    \vspace{-1.2em}
\end{wrapfigure}
Finally, we analyze whether the benefit of ATG is stable across different backbone models. 
As shown in Figure~\ref{fig:atg_backbone_generalization}, ATG consistently outperforms the strongest baseline under each open-source backbone. 
Averaged over the three benchmarks, ATG improves the best baseline from 25.51 to 56.10 on Mistral-7B, from 19.75 to 58.56 on Gemma-7B, and from 29.52 to 62.93 on Llama-3-8B. 
These consistent gains indicate that ATG is not tied to a specific model family. 
Instead, it complements backbone capability by organizing local model decisions into a graph structure. 
Stronger backbones provide better local reasoning, while ATG provides dependency tracking, parallel scheduling, pre-execution validation, and localized recovery. 
This suggests that improving the control framework can be an effective alternative to relying solely on larger backbone models.

\section{Conclusion and Future Work}
We propose \textbf{ATG}, a training-free control framework that organizes LLM-agent planning and execution as an explicit directed task graph. By exposing subtask dependencies, recording graph evolution, enabling dependency-aware execution, and supporting localized repair, ATG improves both success rate and execution efficiency. Experiments on ALFWorld, WebShop, and ScienceWorld show consistent gains across 7B--8B open-source backbones.

\section{Limitation}
\label{limitation}
ATG has several limitations. It depends on the backbone LLM's decomposition ability, and failure localization can be difficult under noisy observations or long-range dependencies. Our experiments focus on text-based benchmarks, while multimodal and real-world settings require further validation. ATG also introduces extra overhead for simple tasks.

\bibliographystyle{unsrtnat}
\bibliography{references}

\end{document}